\documentclass[journal]{IEEEtran}

\usepackage{url}

\usepackage{times}
\usepackage{epsfig}
\usepackage{graphicx}
\usepackage{amsmath}
\usepackage{amssymb}
\usepackage{algorithm,algpseudocode}

\usepackage{bm}
\usepackage{caption}

\usepackage{url}
\usepackage{booktabs}
\usepackage{multirow}
\usepackage{multirow}
\usepackage{wrapfig}
\usepackage{wrapfig,lipsum,booktabs}
\usepackage{graphicx}
\usepackage{array}
\usepackage{color, colortbl}
\usepackage{array,hhline}
\usepackage[]{hyperref}

\usepackage{color, colortbl}

\definecolor{LightCyan}{rgb}{0.88,1,1}

\newif\ifshowfig\showfigtrue

\ifCLASSINFOpdf
\else
\fi
\begin{document}

%\title{Recognition of Emotions in Images Based on Weakly Supervised Emotion Intensity Learning}
\title{RandAlign: A Parameter-Free  Method for Regularizing Graph Convolutional Networks}

%\author{ %To be finalized
\author{Haimin~Zhang,
       % John~Doe,~\IEEEmembership{Fellow,~OSA,}
        and~Min~Xu,~\IEEEmembership{Member, IEEE}

         %<-this % stops a space

 \thanks{Haimin Zhang, Min Xu (\emph{corresponding author})  are with the School of Electrical and Data Engineering, Faculty of Engineering and Information Technology, University of Technology Sydney, 15 Broadway, Ultimo, NSW 2007, Australia (Emails: Haimin.Zhang@uts.edu.au, Min.Xu@uts.edu.au).}% <-this % stops a

}
\maketitle
\pagestyle{empty}
\thispagestyle{empty}


\begin{abstract}

  Studies continually find that message-passing graph convolutional networks suffer from the  over-smoothing issue.
  %This issue can lead to significantly reduced generalization performance when repeatedly applying  graph convolutions.
  Basically, the issue of over-smoothing   refers to the phenomenon that the learned embeddings for all nodes can become very similar to one another and therefore are uninformative after repeatedly applying  message passing iterations.
  %Intuitively, we can expect the generated embeddings become smooth asymptotically layerwisely, that is each layer of graph convolution generates a smoothed version of the input embeddings.
  Intuitively, we can expect the generated embeddings become smooth asymptotically layerwisely, that is each layer of graph convolution generates a smoothed version of  embeddings as compared to that generated by the previous layer.
  %which is a core issue that prevents us from building deep graph  convolutional network models. %restrict the performance of the models.
  %Recent studies show that oversmoothing is one of the major issues that restrict the performance of graph convolutional networks, resulting in the model generalizing poorly on unseen data.
  Based on this intuition, we propose RandAlign, a stochastic regularization  method for  graph convolutional networks.
  %{\color{blue} test}
  %The essential idea of Lrn\&Align consists of two steps.
  The idea of RandAlign is to randomly align the learned embedding for each node with that generated by the previous layer using random interpolation in each  graph convolution layer.
  %Then it aligns a newly generated node embedding  with the original embedding  through reducing the angle between the two embeddings.
  %Through  aligning the learned embeddings,  the over-smoothness of the learned embeddings is explicitly reduced.
  Through  alignment,  the smoothness of the  generated embeddings is explicitly reduced.
  To better maintain the benefit yielded by the graph convolution, in the alignment step we introduce to first scale the  embedding of the previous layer to  the same norm as the generated embedding and then perform random interpolation for aligning the generated embedding.
  %We scale the embedding of the previous layer to have the same norm as the generated embedding and then perform random interpolation.
  RndAlign is a parameter-free method and can be directly applied without introducing additional trainable weights or hyper-parameters.
  We experimentally evaluate RandAlign  on different graph domain tasks  on seven  benchmark datasets.
  The experimental results show that RandAlign is a generic method that improves the generalization performance of various graph convolutional network models and also improves the numerical stability of optimization, advancing the state of the art performance for graph representation learning.

\end{abstract}


\IEEEpeerreviewmaketitle

\iftrue
\section{Introduction}
\label{introduction}
Graph-structured data are very commonly seen in the real world \cite{hamilton2020graph,feng2020graph}.
%Graphs are a ubiquitous structure in the real world \cite{hamilton2020graph,feng2020graph}.
Social networks, protein and drug structures, 3D meshes and  citation networks---all of these types of data can be represented using graphs.
It is of considerable significance to design and develop models that are able to  learn and  generalize from this kind of data. %graph-structured data.
The past years have seen a surge in studies on representation learning on graph-structured data, including techniques for deep graph embedding, graph causal inference and generalizations  of convolutional neural networks to non-Euclidean data \cite{hamilton2020graph}.
These advances have produced new state of the art results in a wide variety of domains, including recommender systems, drug discovery, 2D and 3D computer vision, and question answering systems \cite{sun2020graph,chen2022tinykg,guo2021bilinear,huang2020multitask,deng2022voxel}.

\iffalse
A wide variety of real world data are essentially in a graph structure.
Online social networks, citation networks, protein and drug structures, e-commerce entities---all of these types of data can be represented using graphs.
Designing and developing  models that are able to learn and generalize from graph-structured data is of considerable significance.
Over the past years, increasing effects have been devoted to graph representation learning, including deep graph embedding, generalizations of spatial convolutions to non-Euclidean graph structured data and approaches for message propagation \cite{hamilton2020graph}.
These advances have produced new state of the art results in a variety of tasks, including recommendation systems \cite{he2020lightgcn}, drug discovery \cite{sun2020graph}, community-based question answering \cite{zhang2021graph} and  bioinformatics \cite{zhang2021graphbioinformatics}.
\fi

Unlike images and natural languages, which essentially have a grid or sequence structure, graph-structured data  have an underlying structure in non-Euclidean spaces.
It is a complicated task to develop  models that can generalize over general graphs.
%Unlike images and sequence data,
%Learning on graph-structured data is a complicated task.
%The major challenge in representation learning on graph-structured data is that they have an underlying structure that is in non-Euclidean spaces.
Early attempts \cite{gori2005new,scarselli2008graph} on graph representation learning primarily  use a recursive neural network  which iteratively updates node states and exchanges information until these node states  reach a stable  equilibrium.
%node representations until  reaching a stable point for embedding generations.
Recent years have seen the popularity of graph convolutional networks for graph-structured data.
Graph convolutional networks are derived as a generalization of convolutions to non-Euclidean data \cite{bruna2014spectral}.
The fundamental feature of graph convolutional networks is that it utilizes a  message passing    paradigm  in which messages are exchanged between nodes and updated using  neural networks \cite{gilmer2017neural}.
%Today, graph convolutional networks have become the dominant approach for general graphs.

%In recent years, derived as a generalization of convolutions, the graph convolutional network framework has become the dominant approach over  graph-structured data.
%The defining feature of the graph convolutional network framework is that it uses a form of message passing in which  messages are exchanged between nodes and updated with  neural networks \cite{gilmer2017neural}.
%Recently, graph neural networks have become the dominant paradigm in graph representation learning due to their strong performance.

%The past years have seen considerable progress made message-passing graph neural networks.

This paradigm of message passing  is basically a differentiable variant of belief propagation \cite{dai2016discriminative}.
During each message-passing iteration, the representation for each node is updated according to the information aggregated from the node's neighborhood.
%At each iteration, every node aggregates information from its local neighborhood.
This local feature-aggregation behaviour is analogous to that of the convolutional kernels in convolutional neural networks, which aggregates feature information from spatially-defined patches in an image.
Message passing is the core of current graph convolutional networks, but it also has major drawbacks.
Theoretically, the power of message-passing graph convolutional networks is inherently bounded by the Weisfeiler-Lehman  isomorphism test \cite{xu2019powerful,morris2019weisfeiler}.
Empirically, studies have continually found that
message-passing graph neural networks  suffer from the problem of over-smoothing, and this
issue of over-smoothing can be viewed as a consequence of the neighborhood
aggregation operation \cite{hamilton2020graph}.
%Over-smoothing is a major issue in current graph

%The idea of over-smoothing is that the embeddings for all the nodes in the graph can converge to similar values after many iterations of message passing.

The problem of over-smoothing is that after a number of message passing iterations, the representations for all the nodes in the graph can become very similar to one another.
This is  problematical because node-specific information becomes lost when we add more graph convolutional layers to the model.
This  makes  it difficult to capture long-term dependencies in the graph using deeper layers.
Due to the over-smoothing issue, basic graph convolutional network models such as GCN \cite{kipf2016semi} and GAT \cite{velickovic2018graph} are restricted to a small number of layers, e.g., 2 to 4 \cite{Zhao2020PairNorm}.
Further increasing the number of layers will lead to significantly reduced generalization performance.
This is different from convolutional neural networks, the performance of which  can be considerably  improved by using very deep layers.
Study also shows that the issue of over-smoothing can cause  overfitting  or  underfitting  for different graph domain tasks \cite{zhang2022ssfg}.

Increasing efforts have been devoted to understanding and addressing the over-smoothing problem over the past years.
From the graph signal processing view, applying message passing in a basic graph convolutional network is analogous to applying a low-pass convolutional filter, which produces a smoothed version of the input signal \cite{zhu2020simple}.
Li et al. \cite{li2018deeper} showed that graph convolution  is  a special form of
Laplacian smoothing \cite{taubin1995signal} and proved that repeatedly applying Laplacian smoothing can lead to node representations  becoming indistinguishable from each other. % converging to similar values.
Zhao  et al. \cite{Zhao2020PairNorm} proposed a normalization layer named PariNorm that ensures the total
pairwise feature distance remains unchanged across layers to prevent node features from converging  to similar values.
%Zhang et al. \cite{zhang2022ssfg} showed that over-smoothing can lead to both over-fitting and under-fitting in different domain tasks and introduced to stochastically
Zhang et al. \cite{zhang2022ssfg}  introduced to stochastically scale features and gradients (SSFG) during training.
This method explicitly breaks the norms of generated embeddings becoming over-smoothed for alleviating over-smoothing.

\iffalse
In recent years, increasing efforts have been devoted to understanding and addressing the over-smoothing issue.
Zhao  \emph{et al.} \cite{zhao2019pairnorm} proposed PariNorm, a normalization layer that ensures the total
pairwise feature distance remains to be constant across layers, preventing node features from converging  to similar values.
Rong \emph{et al.} \cite{rong2020dropedge} proposed the DropEdge method as a variant of Dropout  for regularizing graph  networks.
DropEdge randomly removes a number of edges from the input graph at each training epoch.
It works as a data augmentation method and a message passing reducer.
Most of these method either change the model architectures or basically work as a regularization technique/normalization or focus on the aggregation stage of message passing. % ORDERED GNN: ORDERING MESSAGE PASSING TO DEAL WITH HETEROPHILY AND OVER-SMOOTHING
Previous work \cite{zhang2022ssfg} shows that over-smoothing can lead to both over-fitting and under-fitting in model training.
\fi

As introduced above, the learned embeddings for all nodes become very similar to one another when  over-smoothing occurs.
When it comes to becoming very similar to one another, we can understand it from two respects: { (1) These embeddings have a small cosine similarity between one and another;  (2) The norms of these embeddings are close to each other.}
The SSFG method \cite{zhang2022ssfg} is effective through addressing the norms of node embeddings converging to the same value with regard to the second respect.
%The SSFG method explicitly apply a random scaling factor to each embedding to break the norm smoothness for preventing over-smoothing.
However, the issue of node embeddings having a small cosine similarity between one and another is explicitly addressed by the SSFG method.
As aforementioned, the  over-smoothing problem comes after repeatedly message passing iterations.
Intuitively, we can expect the learned embeddings for the nodes become smoothed layerwisely or asymptotically
layerwisely.
That is, each message passing iteration produces a smoothed version of the input embeddings. %makes all node embeddings close to each other.
%While these efforts have been made, over-smoothing is still a key issue in graph convolutional networks.
%It is significant to make further advances to tackle the issue of over-smoothing.
%As introduced above, the issue of over-smoothing comes with repeatedly applying graph convolutions.
%As introduced above, over-smoothing occurs when more graph convolutional layers are added to the model.
%Intuitively each layer of message passing makes the embeddings more over-smoothness than the previous layer.
In this paper we first show, through an  example, the intuition that each layer of graph convolution can make the generated node embeddings closer to each other than  the input embeddings.
%Based on this intuition, we propose random alignment (RandAlign) for reducing over-smoothing in the  embedding generation process.
%Based on this intuition, we propose to randomly align (RandAlign) the generated embedding for each node with that generated by the previous layer for reducing smoothness of the generated embeddings.
Based on this intuition, we propose RandAlign, a stochastic regularization method for  graph convolutional networks.
The idea of  RandAlign is to randomly align the generated embedding for each node with that generated by the previous layer.
Because the embeddings generated by the previous layer are less close to each other, the problem of over-smoothing with regard to the first respect is explicitly reduced through alignment.

In  alignment, we sample a factor from the standard uniform distribution and then align the generated embedding for each node with that generated by the previous layer using convex combination.
%The angles between learned embeddings with those of the previous layer are randomly reduced in alinement.
Therefore our RandAlign method does not introduce additional trainable parameters or hyper-parameters.
%It can be applied to current message-passing graph convolutional networks without the parameter tuning procedure.
It can be applied to current message-passing graph convolutional networks in plug and play manner.
% Lrn\&Align can be used with a wide variety of  graph convolutional models.
We show through experiments that RandAlign is a generic method that improves the generalization performance of a variety of  graph convolutional networks including GCN  \cite{kipf2017semi}, GAT \cite{velickovic2018graph}, GatedGCN \cite{bresson2017residual}, SAN \cite{kreuzer2021rethinking} and GPS \cite{rampasek2022recipe}. % which combines a graph convolutional network with transformer.
%It can help to address both the overfitting issue and the underfitting issue.
We also show that RandAlign is effective on seven popular datasets on different graph domain tasks, including  graph classification and node classification, advancing the state of the art results for graph representation learning on these datasets.
%Thus it is a general method for reducing over-smoothing and improving the model performance. %learning on graph-structured data with graph convolutional networks.
%We experimentally validate our method on different tasks, including graph classification and node classification, on six popular datasets.
%The results show that Lrn\&Align advances the state of the art results on the datasets.

The main contributions of this paper can be summarized as follows:
\begin{itemize}
 \item %We propose RandAlign for regularizing graph convolutional networks.
       We propose a stochastic regularization method named RandAlign for graph convolutional networks.
       %We propose RandAlign, a stochastic regularization method  for graph convolutional networks.
       RandAlign randomly aligns the learned embedding for each node with that learned by the previous layer using random interpolation. % in each layer of graph  convolution.
       This explicitly reduces the smoothness of the generated embeddings.
       %Moreover, in alignment we introduce to scale the embedding of the previous layer to the same norm as the generated embedding before performing random interpolation.
       %Moreover, we introduce a step to scale the embedding of the previous layer to the same norm as the generated embedding and then perform random interpolation for the embedding alignment.
       Moreover,  we introduce to first scale the embedding of the previous layer to the same norm as the generated embedding and then perform random interpolation  for aligning the generated embedding.
       This scaling step helps to maintain the benefit yielded by graph convolution in the aligned embeddings.

  \item RandAlign is a parameter-free method which does not introduce additional trainable  parameters or hyper-parameters. It can be directly applied to current graph convolutional networks  without increasing the model complexity and the  parameter tuning procedure.

  \item We demonstrate that RandAlign is a generic method that consistently improves the generalization performance of various graph convolutional network models,  advancing the state of the art results on different graph domain tasks on seven popular benchmark datasets.
      We also show that RandAlign helps to improve the numerical stability of optimization.

  %\item We propose a method referred to as Lrn\&Align which tackles the issue of over-smoothing  through layerwisely learning and aligning node embeddings. We show that Lrn\&Align is a general method that  improves the performance of a variety of message passing graph convolutional networks.

  %\item Lrn\&Align  is parameter-free method, it can be directly applied on current graph convolutional networks  without  laborious parameter tuning.
  %\item Lrn\&Align has a high generalization performance  on different graph domain tasks, advancing the state of the art results for graph representation learning on a variety of benchmark datasets.
\end{itemize}

\iffalse
\textbf{(1)} we present an parameter-free method that consistently helps reduce oversmoothing, improving the graph representation learning performance on different graph domain tasks; we show that Lrn\&Align helps improve the model performance with increased graph convolutional layers; \textbf{(3)} it is intuitive to understand. it is parameter-free method, it can be applied to easisting GCN models without tuning parameters; \textbf{(2)} We validated our method on a variety of benchmark datasets. Lrn\&Align improves the generalization performance, producing new state of the art results on several datasets.
\fi

\fi

\iftrue
\section{Related Work}
\subsection{Graph Convolutional Networks}

The first-generation graph neural work models were developed by  Gori et al. \cite{gori2005new} and Scarselli et al. \cite{scarselli2008graph}.
These models generalize recursive neural networks for  general graph-structured data.
Motivated by the success of convolutional neural networks for Euclidean data, recent years have seen increasing studies on graph convolutional networks which generalize Euclidean convolutions to the non-Euclidean graph domain.
Current graph convolutional networks can be  categorized into  spectral approaches and  spatial approaches \cite{wu2020comprehensive}. %spectral graph theory

The spectral approaches are based on spectral graph theory.
The key idea in these approaches is that they construct graph convolutions via an extension of the Fourier transform to graphs, and a full model is defined by stacking multiple graph convolutional layers.
For example, Bruna et al. \cite{bruna2014spectral} proposed to construct graph  convolutions based on the eigendecomposition of the graph Laplacian.
%The graph convolutions are constructed via an extension of the Fourier transform to graphs.
Following on Bruna's work, Defferrard et al. \cite{defferrard2016convolutional} introduced to construct convolutions based on the Chebyshev expansion of the graph Laplacian.
This approach eliminates the process for graph Laplacian decomposition and results in spatially
localized filters.
Kipf and Welling \cite{kipf2017semi} simplified the previous methods by introducing the popular GCN architecture, wherein the filters are defined on the 1-hop neighbourhood as well as the node itself.

Unlike the spectral approaches, the spatial  approaches directly define convolutions on the graph and  generate node embeddings  nodes by aggregating  information from a local neighbourhood.
Monti et al. \cite{monti2017geometric} proposed a mixture model network, referred to as MoNet, which is a spatial approach that generalizes convolutional neural network architectures to graphs and manifolds.
Velickovic et al. \cite{velickovic2018graph} introduced to integrate the self-attention mechanism which assigns an attention weight or importance value to each neighbour in local feature aggregation into graph convolutional network models.
Bresson et al. \cite{bresson2017residual} proposed residual gated graph convnets, integrating edge gates, residual connections \cite{he2016deep} and batch normalization \cite{ioffe2015batch} into the graph convolutional neural network model.
Balcilar et al. \cite{balcilar2021analyzing} demonstrated that both spectral and spatial graph convolutional networks are essentially message passing neural networks that use a form of message passing for node embedding generation.

\begin{figure*}[ht]
\vskip 0.2in
\begin{center}
\centerline{\includegraphics[width=0.85\textwidth]{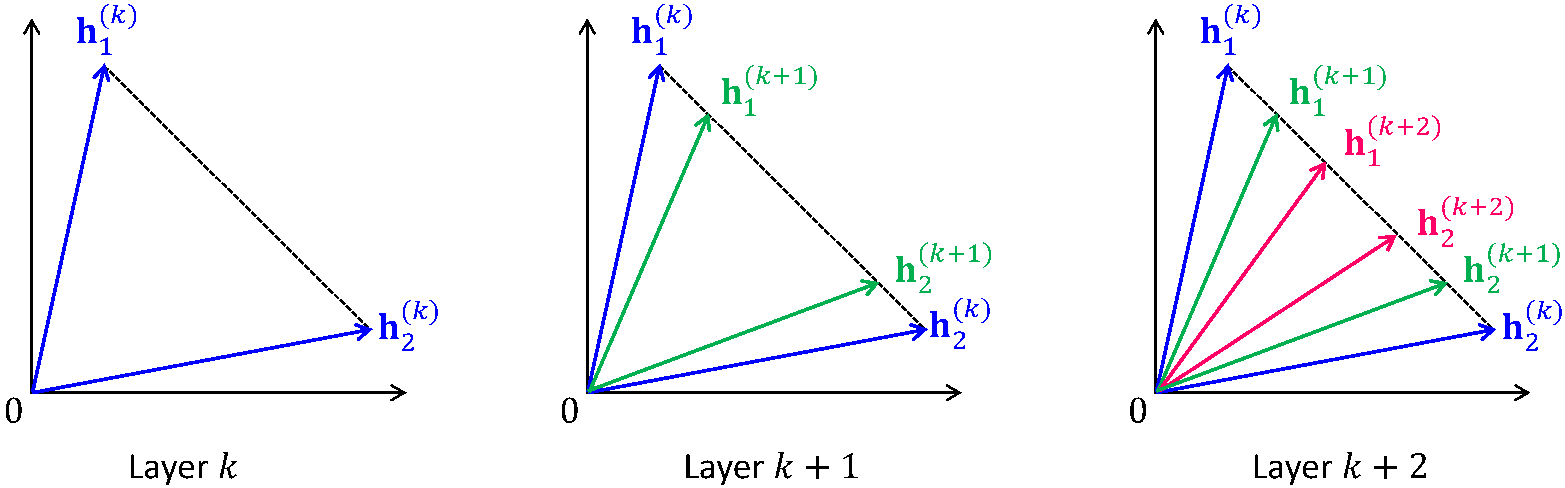}}
\caption{An illustrative  example for understanding the over-smoothing issue. We consider a two node fully connected graph and use a GAT model that layerwisely learn embeddings using the equation $\mathbf{h}_{i}^{(k)} =  \sum_{v \in \mathcal{N}(u)} \alpha_{u,v}   \mathbf{h}_{v}^{(k-1)}$, wherein $\alpha_{u,v}>0$ and $\sum_{v \in \mathcal{N}(u)} \alpha_{u,v}=1$. We have simplified the
model by removing the non-linearity and learnable parameter matrix. We show that the learned embeddings layerwisely become smoothed than the previous layer due to the convex combination of neighbourhood features.}
\label{fig:demon_oversmoothing}
\end{center}
\vskip -0.2in
\end{figure*}

\subsection{The Over-smoothing Problem}

Over-smoothing is a common issue with current graph convolutional neural networks.
Intuitively, this phenomenon of over-smoothing occurs when the information aggregated from the local neighbours starts to dominate the updated node embeddings.
Therefore, a straightforward  way to reduce over-smoothing is to use feature concatenations or skip connections \cite{hamilton2020graph}, which are commonly used in computer vision to build deep convolutional network architectures.
Feature concatenations and skip connections can preserve  information learned by previous graph convolutional layers. %rounds of message passing during the update step. % that bear strong analogy to techniques used in computer vision to build deeper CNN architectures
%was proposed in the GraphSAGE framework, which was one of the rst works to highlight the possible benets of these kinds of modications to the update function [Hamilton et al., 2017a].
%Drawn inspiration from the gating methods in recurrent neural networks, researchers also proposed gated updates in aggregating information from neighbours \cite{li2015gated,bresson2017residual}.
Inspired by the gating methods used to improve recurrent neural networks, researchers also proposed gated updates in aggregating information from local neighbours \cite{li2015gated,bresson2017residual}.
These gated updates are very effective in building deep graph convolutional network architectures, e.g., 10 or more  layers.
Zhao et al. \cite{Zhao2020PairNorm} proposed the PairNorm method  to tackle  oversmoothing by ensuring the total pairwise feature distance across layers to be constant.
%Chen \emph{et al.} \cite{chen2020measuring} introduced to add a MADGap-based regularizer  and  use adaptive edge optimization  to prevent oversmoothing.
Zhang et al. \cite{zhang2022ssfg} proposed a stochastical regularization method called SSFG that randomly scales features and gradients in the training procedure.
Empirically, SSFG can help to address both the overfitting issue and the underfitting issue for different graph domain tasks.
Chen et al. \cite{chen2022optimization}  proposed a graph convolution operation, referred to as graph implicit nonlinear diffusion, that can  implicitly have access to infinite hops of neighbours while adaptively aggregating features with nonlinear diffusion to alleviate the over-smoothing problem.

\fi

\iftrue
\section{Methodology}
\label{methodology}

In this section, we begin by introducing the notations and the message passing framework.
Then we introduce the over-smoothing issue with  graph convolutional networks.
Finally we describe the proposed RandAlign method for regularizing graph convolutional networks through reducing the over-smoothing problem.

\subsection{Preliminaries}
Formally, a graph $G = (V, E)$ can be defined by a set of nodes, or called vertices, $V$ and a set of edges  $E$ between these nodes.
%$n=|\mathcal{V}|$ is the number of nodes in $\mathcal{G}$.
An edge going from node $u\in{V}$ to node $v\in{V}$ is denoted as $(u,v)$.
Conveniently, the graph $G$ can be represented using an adjacent matrix $\mathbf{A} \in \mathbb{R}^{|V| \times |V|}$, in which $\mathbf{A}_{u,v}=1$ if $(u,v)\in {E}$ or $\mathbf{A}_{u,v}=0$ otherwise.
The degree matrix  of ${G}$ is a diagonal matrix and is denoted as $\mathbf{D}\in \mathbb{R}^{|V| \times |V|}$, in which $\mathbf{D}_{ii}=\sum_j \mathbf{A}_{ij}$.
%$\mathbf{X}\in \mathbb{R}^{N\times d}$ is the attribute or feature matrix associated  the nodes, $\mathbf{x}_v \in \mathbb{R}^d$ is the attribute vector for node $v$.
%$\{\mathbf{x}_v, \forall v \in \mathcal{V}\}$ is the set of features or attributes associated with the nodes.
%$\mathbf{X} \in \mathbb{R}^{|V| \times m}$ is the node-level feature or attribute associated with the graph.
%The node-level feature or attribute associated with $u\in V$ is denoted as $\mathbf{x}_u \in \mathbb{R}^{m}$, where $m$ is the dimension of the node features.
The node-level feature or attribute associated with $u\in V$ is denoted as $\mathbf{x}_u$.
The graph Laplacian is defined as  $\mathbf{L}=\mathbf{D}-\mathbf{A}$, and the symmetric normalized Laplacian is defined as $\mathbf{A}_{sym} = \mathbf{I}_n -\mathbf{D}^{-1/2}\mathbf{A}\mathbf{D}^{-1/2}$, where $\mathbf{I}_n$ is a $|V| \times |V|$ identity matrix.

Message passing is at the core of current graph convolutional networks.
In the message passing paradigm, nodes aggregated message from neighbours and updated their embeddings according to the aggregated information in an iterative manner.
%The message passing  paradigm utilizes  a type of neural message passing, in which messages are exchanged between nodes and updated with neural networks  \cite{gilmer2017neural}.
This message passing update can be  expressed as follows \cite{hamilton2020graph}:
\begin{equation} \label{eq:general_messagepassing}
    \begin{aligned}
     \mathbf{h}_{u}^{(k)} = f^{(k)}\left(\mathbf{h}_u^{(k-1)}, agg^{(k)}(\{\mathbf{h}_v^{(k-1)}, \forall v \in \mathcal{N}(u)\}) \right),
    \end{aligned}
\end{equation}
where $f$ and $agg$  are  neural networks, and $\mathcal{N}(u)$ is the set  of $u$'s neighbouring nodes.
The superscripts are used for distinguishing the embeddings and functions at different iterations.
During each  message-passing iteration, a hidden representation $\mathbf{h}_u^{(k)}$ for each node $u\in V$ is updated according to the  message aggregated from $v$'s  neighbouring nodes.
The embeddings at $k=0$ are initialized to the node-level features, i.e., $\mathbf{h}_u^{(0)}=\mathbf{x}_u, \forall u\in V$.
After $k$ iterations of message passing, every node embedding contains information about its $k$-hop
neighborhood.

\subsection{The Over-smoothing Problem}
While message passing is at the heart of  current graph convolutional networks, this  paradigm also has major bottlenecks.
%Studies have continually found that message-passing graph convolutional networks suffer from the  over-smoothing issue.
Studies  continually show that over-smoothing is a common issue with current message-passing graph convolutional networks.
%Over-smoothing is core limitation in current graph convolutional networks, and this issue an be viewed as a consequence of the neighborhood aggregation operation, which is at the heart of the message passing paradigm.
%A major drawback networks is the issue over-smoothing, which can be viewed as a consequence of the neighborhood aggregation operation.
The intuitive idea of over-smoothing is that after repeatedly applying  message passing, the representations for all nodes in the graph  can become very similar to one another, therefore node-specific features become lost.
Due to this issue, it is impossible to build deeper models to capture the longer-term dependencies in the graph.

From the perspective of graph signal processing, the graph convolution of the GCN model \cite{kipf2016semi} can be seen as a special form of Laplacian smoothing \cite{li2018deeper} that basically updates the  embedding for a node using the weighted average of the node's itself and its neighbour embeddings.
But after applying too many rounds of Laplacian smoothing, the representations for all nodes will become indistinguishable from each other.

\iffalse
From the view of graph signal processing, the graph convolution of the GCN model \cite{kipf2016semi} is a special form of Laplacian smoothing \cite{li2018deeper}, which makes the generated embeddings for  nodes in the same cluster become more similar.
But applying too many Laplacian smoothing, the representations for all nodes will become indistinguishable from each other.
\fi

Formally, the issue of over-smoothing  can be described  through defining the influence of each node's input feature on the final layer embedding of all the other nodes in the graph.
For any pair of node $u$ and node $v$, the influence of node $u$ on node $v$ in a graph convolutional network model can be quantified by examining the magnitude of the corresponding Jacobian matrix \cite{xu2018representation} as follows:
\begin{equation} \label{eq:Jacobian-Xu}
    \begin{aligned}
      %\lim_{k \to +\infty} \tilde{A}_{sym}^{k} \cdot X_{\cdot j} = \pi_j, \hspace{5pt} j=1,2,...,N. \\
      I_{K}(u,v) = \mathbf{1}^{\top} \left(  \frac{\partial \mathbf{h}_v^{(K)}}{\partial \mathbf{h}_u^{(0)}} \right) \mathbf{1},
     \end{aligned}
\end{equation}
where $\mathbf{1}$ is a vector of ones.
$I_K(u,v)$, which is the sum of the entries in the Jacobian matrix $\frac{\partial \mathbf{h}_v^{(K)}}{\partial \mathbf{h}_u^{(0)}}$, is a  measure of how much the initial
embedding of node $u$ influences the final embedding of node $v$.
Given the above definition of influence, Xu et al. \cite{xu2018representation} prove the following theorem:

\textbf{Theorem 1}. For any graph convolutional network model which uses a self-loop update approach and
an aggregation function of the following form:
\begin{equation} \label{eq:oversmoothing-Xu}
    \begin{aligned}
      %\lim_{k \to +\infty} \tilde{A}_{sym}^{k} \cdot X_{\cdot j} = \pi_j, \hspace{5pt} j=1,2,...,N. \\
      agg(\{ \mathbf{h}_v, \forall v\in \mathcal{N}(u) \cup \{u\} \})  = \hspace{70pt} \\
       \frac{1}{g_n(\left| \mathcal{N}(u) \cup \{u\} \right|)}  \sum_{v\in \mathcal{N}(u) \cup \{u\}} \mathbf{h}_v,
     \end{aligned}
\end{equation}
where $g_n$ a normalization function, we have the following:
\begin{equation} \label{eq:oversmoothing-Xu}
    \begin{aligned}
      I_K(u,v) \propto p_{G,K}(u|v),
     \end{aligned}
\end{equation}
where $p_{G,K}(u|v)$ denotes the probability of visiting node $v$ on a length of $K$ random walk starting from node $u$.

Theorem 1 states that when we are using a $K$-layer graph convolutional network model, the influence
of node $u$ on node $v$ is proportional to the probability of reaching node $v$ on a $K$-step random walk starting from node $u$.
The  consequence of this is that as $K\to \infty$ the influence of every node
approaches the stationary distribution of random walks over the graph, therefore the information from local neighborhood is lost.
Theorem 1 applies directly to the models that use a self-loop update approach, but the result can also be generalized  asymptotically for the models that use the basic message passing update in Equation \ref{eq:general_messagepassing}.

\iffalse
% where nodes aggregate messages from neighbors and then update their representations in an iterative fashion

%Over-smoothing  Its main cause is the stacked aggregation layer using the same adjacency matrix \cite{shi2022revisiting}.
%While graph networks have achieved state-of-the-art performance for various  graph-based tasks, these models are mostly restricted to shallow layers, \emph{e.g.,} 2 to 4.
%The success appears to be limited to shallow
Oversmoothing is one of the major issues that restrict the depth of graph networks.
Oversmoothing comes with the nature of graph convolutions which update node features by aggregating neighborhood information.
%Graph networks update node features by aggregating neighborhood information.
Repeatedly applying graph convolutions results in node features across different categories converging to similar values regardless of input features, eliminating the discriminative information from these features.
\fi

\subsection{Proposed Method: RandAlign Regularization}
Over-smoothing is a common issue in message-passing graph convolutional networks.
This issue occurs when the generated node embeddings become over-smoothed and therefore uninformative after repeatedly applying message passing iterations.
This is problematic because  information from local neighbourhood becomes lost when more layers of message passing are added to the model.
Due to the over-smoothing issue, it is difficult to stack  deeper graph convolutional layers to capture long-term dependencies of the graph.

%This issue occurs when the generated node embeddings become over-smoothed after repeatedly applying layers of message passing.
%This is problematic because the information from the input node features becomes lost.

%When more layers of message passing are added to the model, the generated node embeddings become over-smoothed and uninformative, and therefore the model performance can be significantly reduced.
%This is problematic  because it makes  graph convolutional network models can only capture limited dependencies of the graph.

\iffalse
The consequence of over-smoothing is that when we are using more layers of message passing, the generated node embeddings become over-smoothed and uninformative, and therefore the information from the input node features becomes lost.
This is problematic  because it makes  graph convolutional network models can only capture limited dependencies of the graph.
\fi

As introduced in the introduction section, when it comes to   embeddings becoming over-smoothed, we can understand it in two respects:  (1) these embeddings have a small cosine similarity between one and another; and (2) the norms of these embeddings are close to each other.
The SSFG method \cite{zhang2022ssfg} stochastically scales the norms of the learned embeddings at each layer during training.
This method explicitly breaks the norms of the embeddings converging to the similar value regarding the second respect for reducing over-smoothing.
With regard to the first respect, however, the  issue of node embeddings having a small cosine similarity between one and another is not explicitly addressed.

As discussed above, the issue of over-smoothing occurs  after applying too many layers of message passing.
%Theorem 1 provides theoretical analysis of the issue of over-smoothing.
Intuitively, we can expect the learned  embeddings for all nodes in the graph become smoothed layerwisely or asymptotically layerwisely.
Based on this intuition, we propose RandAlign which randomly aligns the learned embedding for each node with that generated by the previous layer for regularizing graph convolutional networks.
%From the perspective of graph signal processing, the graph convolution of the basic GCN model \cite{kipf2017semi} is a low-pass filter \cite{balcilar2021analyzing}, and it smooths the signal on the graph.
Here, we first show an example to demonstrate the intuition that each layer of graph convolution produces a smoothed version of the input.
%We show this through an illustrative example.
Consider we have a two node fully connected graph and use the GAT model \cite{velickovic2018graph} to generate node embeddings (see Figure \ref{fig:demon_oversmoothing}).
The GAT model updates the embedding for node $u$ at the $k$-th layer of message passing using a weighted sum of information from its neighbours as follows:

\begin{equation} \label{eq:loss_int}
    \begin{aligned}
    %\tilde{ x} = \bm{x} + \Pi_{ \bm{x}+\mathcal{C} }( \beta \sum_{i=1}^{N} \eta_i (\bm{x})),
    % we have simplied the model by removing the non-linearity
    %h_{i}^{l+1} = \sigma \left( \sum_{v' \in \mathcal{N}(u)} \alpha_{u,v}  \mathbf{W}^{(k)}  \mathbf{h}_{v'}^{(k)} \right),
    \mathbf{h}_{u}^{(k)} =  \sum_{v \in \mathcal{N}(u)} \alpha_{u,v}^{(k)}  \mathbf{W}^{(k)}  \mathbf{h}_{v}^{(k-1)},
    \end{aligned}
\end{equation}
where $\alpha_{u,v}^{(k)}$ is the attention weight on neighbour $v\in \mathcal{N}(u)$ when aggregating
information at node $u$, and $\mathbf{W}^{(k)}$ is a learnable weight matrix.
Note that we have simplified the model by removing the non-linearity as compared to the original GAT \cite{velickovic2018graph}.
%We have simplified the model by removing the non-linearity as compared to the original GAT \cite{velickovic2018graph}.
%In the  equation above, $\alpha_{u,v}$ denotes the attention weight on neighbor $v\in \mathcal{N}(u)$ when aggregating information at node $u$, and $\mathbf{W}^{(k)}$ is a learnable parameter matrix.
%The attention weights are defined using the softmax function as:
The attention weight $\alpha^{(k)}_{u,v}$ is defined using the softmax function as follows:
\begin{equation} \label{eq:loss_int}
    \begin{aligned}
    %\tilde{ x} = \bm{x} + \Pi_{ \bm{x}+\mathcal{C} }( \beta \sum_{i=1}^{N} \eta_i (\bm{x})),
    %h_{i}^{l+1} = \sigma \left( \sum_{v' \in \mathcal{N}(u)} \alpha_{u,v}  \mathbf{W}^{(k)}  \mathbf{h}_{v'}^{(k)} \right), \\
    \alpha_{u,v}^{(k)} = \frac{\text{exp}\left( {\mathbf{a}^{(k)}}^{\top} \left[ \mathbf{W}^{(k)} \mathbf{h}^{(k-1)}_u \parallel \mathbf{W}^{(k)} \mathbf{h}^{(k-1)}_v \right] \right)}{\sum_{v'\in\mathcal{N}(u)} \text{exp}\left( {\mathbf{a}^{(k)}}^{\top} \left[ \mathbf{W}^{(k)} \mathbf{h}^{(k-1)}_u \parallel \mathbf{W}^{(k)} \mathbf{h}^{(k-1)}_{v'} \right] \right) },
    \end{aligned}
\end{equation}
where $\mathbf{a}^{(k)}$ is learnable vector, and $\parallel$ denotes the concatenation operator.
With the softmax function, the attention weights are normalized to 1, i.e., $\sum_v\alpha_{u,v}=1$.
Therefore, the learned embedding $\mathbf{h}_{u}^{(k)} $ is essentially a convex combination of the information from $u$'s neighbours.
As shown in Figure \ref{fig:demon_oversmoothing},
$\mathbf{h}_1^{(k+1)}$ and $\mathbf{h}_2^{(k+1)}$ are on the dash line between $\mathbf{h}_1^{(k)}$ and $\mathbf{h}_2^{(k)}$, and $\mathbf{h}_1^{(k+2)}$ and $\mathbf{h}_2^{(k+2)}$ are on the dash line between $\mathbf{h}_1^{(k+1)}$ and $\mathbf{h}_2^{(k+1)}$.
Thus, each layer of the message passing  makes the generated embeddings  more smoothed than those generated by the previous layer.
%After each layer of graph convolution, the generated embeddings are more smoothed than the previous layers.
%As more layers are added to the model, the learned embeddings become over-smoothed and thus the  information from local neighbours become lost.
As more message passing iterations are applied, the learned embeddings  become indistinguishable from each other and thus the  information from local neighbours become lost.

When the embeddings become smoothed, the average cosine similarity between one and another is reduced compared to  that of the embedddings generated by the previous layer.
%An important consequence of the embeddings becoming smoothed layerwisely is that the average angle between pairs of the learned embeddings is reduced compared to  that of the previous layer.
As shown in Figure \ref{fig:demon_oversmoothing}, the cosine between $\mathbf{h}_1^{(k+1)}$ and $\mathbf{h}_2^{(k+1)}$ is small as compared to the cosine between $\mathbf{h}_1^{(k)}$ and $\mathbf{h}_2^{(k)}$, and the cosine between $\mathbf{h}_1^{(k+2)}$ and $\mathbf{h}_2^{(k+2)}$ is small as compared to the cosine between $\mathbf{h}_1^{(k+1)}$ and $\mathbf{h}_2^{(k+1)}$.
To reduce the smoothness of the generated embeddings, we randomly align the generated embedding for each node with that generated by the previous layer.
%Based on the above  analysis, we propose Lrn\&Align that layerwisely aligns the learned embeddings with those of the previous layer to reduce the over-smoothing problem.
Specifically, in each  layer we first apply the message passing in Equation (\ref{eq:general_messagepassing}) to generate an intermediate embedding for each node $u\in V$:
\begin{equation} \label{eq:intermediate_embedding}
    \begin{aligned}
     \overline{\mathbf{h}}_{u}^{(k)} = f^{(k)}\left(\mathbf{h}_u^{(k-1)}, agg^{(k)}(\{\mathbf{h}_v^{(k-1)}, \forall v \in \mathcal{N}(u)\}) \right).
    \end{aligned}
\end{equation}
Then we align $\overline{\mathbf{h}}_{u}^{(k)}$ with $\mathbf{h}_{u}^{(k-1)}$ using random interpolation.
%We adopt a random alignment approach.
%To better maintain the representational ability of  embedding $\overline{\mathbf{h}}_{u}^{(k)}$,  we first rescale $\mathbf{h}_u^{(k-1)}$ to have the same norm as $\overline{\mathbf{h}}_u^{(k)}$, then we apply a random interpolation between the two embeddings. % obtaining the following aligned embedding:
To better maintain the benefit yielded by message passing in the aligned embedding,  we first rescale $\mathbf{h}_u^{(k-1)}$ to have the same norm as $\overline{\mathbf{h}}_u^{(k)}$, then we apply a random interpolation between the two embeddings. % obtaining the following aligned embedding:
Finally, the  embedding for node $u \in V$ is updated with the residual connection \cite{he2016deep} as follows:

\begin{equation} \label{eq:rand_alignment}
    \begin{aligned}
    \mathbf{h}_u^{(k)} &= \mathbf{h}_u^{(k-1)} + align(\mathbf{h}_u^{(k-1)}, \overline{\mathbf{h}}_u^{(k)}) \\
     &= \mathbf{h}_u^{(k-1)} + \lambda  \frac{\mathbf{h}_u^{(k-1)}}{\lVert \mathbf{h}_u^{(k-1)}\rVert}  {\lVert \overline{\mathbf{h}}}_u^{(k)}\rVert + (1-\lambda) \overline{\mathbf{h}}_u^{(k)},
    \end{aligned}
\end{equation}
where $align$ is a function for aligning $\overline{\mathbf{h}}_u^{(k)}$  with $\mathbf{h}_u^{(k-1)}$, and  $\lambda \sim U(0,1)$ is sampled from the standard uniform distribution.
By this way, we can keep the representation ability yielded by message passing while reducing the smoothness in the aligned embeddings.
%By this way, we can maintain the representational ability of $\overline{\mathbf{h}}_v^{(k)}$ in the aligned embedding while reducing the angle between the two embeddings.
%By rescaling $\mathbf{h}_v^{(k)}$ to have the same norm as $\overline{\mathbf{h}}_v^{(k+1)}$, we can keep the representational ability of $\overline{\mathbf{h}}_v^{(k+1)}$ in the
%At test time, $\lambda$ to set to a fixed value of 0.5.
Because the expected value of $\lambda$ is 0.5, i.e., $E[\lambda]=0.5$, at test time $\lambda$ is set to a fixed value of $0.5$.
Algorithm \ref{alg:lrnalign} shows the embedding generation algorithm with the message-passing framework and our RandAlign regularization method.

\begin{table*}[th]
%\scriptsize
\caption{Details of the seven benchmark datasets used in the experiments.   } % On COLLAB, edges that represent collaborations up to 2017 are used for training, and edges that represent collaborations in 2018 and 2019 are used for validation and testing, respectively.
\vskip 0.15in
%\small
\centering  % 表居中
\begin{tabular}{ l|c|cccccc }
\toprule
Dataset  &Graphs &Avg. Nodes/graph &\#Training &\#Validation &\#Test &\#Categories & Task\\
%\cmidrule(r){3-5} \cmidrule(r){6-8}
%\cline{2-4} \cline{5-7}
%                          & Mixup  & IAT    & Mixup  & IAT \\
%\midrule[.6pt]

\midrule[.5pt]

MNIST   &70K &40-75  &55,000 &5000 &10,000 & 10 &\multirow{2}{*}{Superpixel graph classification}\\
CIFAR10 &60K &85-150 &45,000 &5000 &10,000 & 10 &\\
\midrule[.5pt]
PascalVOC-SP & 11,355 &479.40 & 8,489 &1,428 &1,429 &20 & Superpixel graph classification \\

\midrule[.5pt]

PATTERN & 14K &44-188 &10,000 &2000 &2000 & 2 &\multirow{2}{*}{Node classification}\\
CLASTER & 12K &41-190 &10,000 &1000 &1000 & 6 & \\
\midrule[.5pt]

Peptides-Func &15,535 &150.90 &70\% &15\% &15\% &10 &Multi-label graph classification\\

\midrule[.5pt]

OGBG-Molhiv &41,127 &25.50 &80\%  &10\%  &10\%  &2 &Binary graph classification \\

\bottomrule

\end{tabular}

\label{table:datasets}
\end{table*}

The proposed RandAlign method is  straightforward to understand.
By  aligning the learned embeddings with those generated by the previous layer, the smoothness of these learned embeddings is explicitly reduced, therefore the overall model performance is improved. %se improving the model generalization performance.
Because the embeddings before alignment are learned using the basic message-passing framework, our RandAlign is a general method that can be applied in different message passing graph convolutional network models to alleviate the over-smoothing problem.
Moreover, our RandAlign method does not introduce additional hyper-parameters or trainable weights, it can be directly applied in a plug and play manner and without the time-consuming hyper-parameter tuning procedure.

\begin{algorithm}[tb]
\caption{The embedding generation process with the message-passing framework and our RandAlign regularization method.}
\label{alg:lrnalign}
\hspace*{0.02in} {\bf Input:} Graph $G =(V, E)$; number of graph convolutional layers $K$; input node features $\{ \mathbf{x}_v, \forall u\in V \}$ \\
\hspace*{0.02in} {\bf Output:} Node embeddings $\mathbf{h}_u^{(K)}$ for all $u\in V$
\begin{algorithmic}[1]
\State {$\mathbf{h}_u^{(0)} \leftarrow \mathbf{x}_u, \forall u\in V $}
\For{$k=1,...,K$}
    \For{$u\in \mathcal{V}$}
       \State $\overline{\mathbf{h}}_{u}^{(k)} = f^{(k)}\left(\mathbf{h}_u^{(k-1)}, agg^{(k)}(\{\mathbf{h}_v^{(k-1)}, \forall v \in \mathcal{N}(u)\}) \right)$  {\color{blue}// generate an intermediate embedding for $u$ using a general message passing model (see Equation (\ref{eq:intermediate_embedding}))}.
    \EndFor

    \For{$v\in \mathcal{V}$}

       \iftrue
        \If{model.training ==  True}
           \State $\lambda \sim U(0,1)$ %{\color{blue} // sample $\lambda$ from the standard distribution during training.}
        \Else \State $\lambda=0.5$
         \EndIf
         \fi

        \State $\mathbf{h}_u^{(k)}=  \mathbf{h}_u^{(k-1)} + \lambda \cdot \frac{\mathbf{h}_u^{(k-1)}}{\lVert \mathbf{h}_u^{(k-1)}\rVert} \cdot {\lVert \overline{\mathbf{h}}_u^{(k)}}\rVert + (1-\lambda) \cdot\overline{\mathbf{h}}_u^{(k)}$  {\color{blue}// update the embedding for $u$ as sum of the aligned embedding and the input node embedding (see Equation (\ref{eq:rand_alignment})).}

        %{\color{blue}// align the intermediate embedding $\overline{\mathbf{h}}_u^{(k)}$ with  $\mathbf{h}_u^{(k-1)}$ (see Equation (\ref{eq:rand_alignment})).}
       \EndFor

\EndFor

\iffalse
\State Initialize a population of particles with random values positions
       and velocities from \textit{D} dimensions in the search space
\While{Termination condition not reached}
\For{Each particle $i$}
    \State Adapt velocity of the particle using Equation \ref{eq:1}
    \State Update the position of the particle using Equation \ref{eq:2}
    \State Evaluate the fitness {$f(\overrightarrow{X}_i)$}
    \If{$f(\overrightarrow{X}_i)<f(\overrightarrow{P}_i)$}
       \State $\overrightarrow{P}_i \gets \overrightarrow{X}_i$
    \EndIf
    \If{$f(\overrightarrow{X}_i)<f(\overrightarrow{P}_g)$}
       \State $\overrightarrow{P}_g \gets \overrightarrow{X}_i$
    \EndIf
\EndFor
\EndWhile
\fi
\end{algorithmic}
\end{algorithm}

%\section{Proposed Method:}

\fi

\section{Experiments}
\label{sec:experments}

\subsection{Datasets and Setup}

\textbf{Datasets}.
The proposed RandAlign  method is evaluated on four graph domain tasks: graph classification, node classification, multi-label graph classification and binary graph classification.
The experiments are conducted on seven benchmark datasets, which are briefly introduced as follows.

\iffalse
%We conduct extensive experiments on the following tasks
%Our experiments are conducted on five recently released  benchmark datasets \cite{dwivedi2020benchmarking}, \emph{i.e.,} MNIST, CIFAR10, PATTERN, CLUSTER and ZINC.
We conduct extensive experiments on different graph domain tasks on five large datasets including  MNIST, CIFAR10, PATTERN, CLUSTER and ZINC \cite{dwivedi2020benchmarking}, and three small datasets including ENZYMES, PROTEINS \cite{morris2020tudataset} and OGBG-MOLTOX21 \cite{hu2020open}.
\fi

\begin{itemize}
  \item  \textbf{MNIST and CIFAR10} \cite{dwivedi2020benchmarking} are two datasets  for superpixel graph classification.
      The original images in MNIST \cite{lecun1998gradient} and CIFAR10 \cite{krizhevsky2009learning} are converted to  superpixel graphs using the SLIC technique \cite{achanta2012slic}.
      Each superpixel represents a small region of homogeneous intensity in the original image.
      %Fig. \ref{fig:samples-mnist-cifar} demonstrates examples of the superpixel graphs in the two datasets.
      %The two datasets are used for evaluation on the graph classification task.
  \item \textbf{PascalVOC-SP} \cite{dwivedi2022long} is also a dataset for superpixel graph classification. There are 11,355 graphs with a total of 5.4 million nodes in PascalVOC-SP. Each superpixel graph  corresponds to an image in Pascal VOC 2011.
     The superpixel graphs in PascalVOC-SP are much large  compared to those in MNIST and CIFAR10 \cite{dwivedi2020benchmarking}.

%Compared to MNIST and CIFAR10, PascalVOC-SP has a large average nodes per graph.
%The graphs prepared after the superpixels extraction have on average 479.40 nodes
   %PascalVOC-SP is a node classification dataset, based on the Pascal VOC 2011 image dataset [18], where each node corresponds to a region of the image belonging to a particular class.

  \item \textbf{PATTERN and CLUSTER} \cite{dwivedi2020benchmarking}. The two datasets are used for inductive node classification.
     The graphs in the two datasets are generated using the stochastic block model \cite{abbe2017community}.
     %We evaluate our model for recognizing specific predetermined subgraphs on PATTERN and for identifying community clusters in the semi-supervised setting on CLUSTER.
     PATTERN is used for evaluating the model for recognizing specific predetermined subgraphs, and CLUSTER is used for identifying community clusters in the semi-supervised setting.

     %PATTERN tests the fundamental graph task of recognizing specific predetermined subgraphs (as proposed in [75]) and CLUSTER aims at identifying community clusters in a semi-supervised setting [44].
     %, which is widely used for modelling communities in social networks through  modulating the intra- and extra-community connections.

  \item \textbf{Peptides-func} \cite{dwivedi2022long} is a dataset of peptides molecular graphs. % which contains of 15,535 graphs with a total of 2.3 million nodes.
      The nodes in the graphs represent  heavy (non-hydrogen) atoms of the peptides, and the edges represent the bonds between these atoms.
      The graphs are categorized into 10 classes based on the peptide functions, e.g., antibacterial, antiviral, cell-cell communication.
      This dataset is used for evaluating the model for multi-label graph classification.
      %We evaluate the model for multi-label graph classification using this dataset.

  \item \textbf{OGBG-molhiv} is a molecule graph dataset introduced in  the open graph benchmark (OGB) \cite{hu2020open}.
      The nodes and edges in the graphs represent atoms and the chemical bonds between these atoms.
      This dataset is used for evaluating the model's  ability  to predict if or not the molecule can inhibit HIV virus replication, which is a binary class classification task.

      %The task on this dataset is binary class classification. That is we evaluate the model's ability  to predict if the molecule can inhibit HIV virus replication.

\end{itemize}

%Dwivedi \emph{et al.} \cite{dwivedi2020benchmarking} recently released new benchmark datasets for evaluating GNNs.
%These datasets are used for four graph-based tasks: node classification (PATTERN, CLUSTER), graph classification (MNIST, CIFAR10), and graph regression (ZINC).
More details of the six datasets, including the dataset sizes and splits, are reported in Table \ref{table:datasets}.

%\subsubsection{Evaluation Metrics}
%\textbf{Evaluation Metrics}.
\textbf{Evaluation Metrics}.
Following  Dwivedi et al. \cite{dwivedi2020benchmarking} and Rampasek et al. \cite{rampasek2022recipe}, the following  metrics are used for different domain tasks.
For the node classification task, the performance is measured using the weighted accuracy.
For superpixel graph classification, we report the classification accuracy on  test set.
For multi-label graph classification on Peptides-func, the performance is measured using average precision (AP) across the categories.
For the binary classification task on OGBG-molhiv, the performance is measured using the area under the receiver operating characteristic curve (ROC-AUC). %

\textbf{Implementation Details}.
We closely follow the experimental setup as Dwivedi et al. \cite{dwivedi2020benchmarking} and Rampasek et al. \cite{rampasek2022recipe} for training the models.
We use the same train/validation/test split of each dataset and report the  mean and standard deviation over 10 runs.
For experiments on MNIST, CIFAR10, PATTERN and CLUSTER, the Adam algorithm \cite{kingma2014adam} is used for optimizing  the models.
The learning rate is initialized to $10^{-3}$ and reduced by a factor of 2 if the loss has not improved for a number of epochs (10, 20 or 30).
The training procedure is terminated when the learning rate is reduced to smaller than $10^{-6}$.
For experiments on PascalVOC-SP, Peptides-func and OGBG-molhiv, the AdamW algorithm \cite{loshchilov2017decoupled} with cosine learning rate schedule is used for training  the models.
The training epochs are set to 300, 200 and 150, respectively.

\iffalse
For fair comparison, we use the same train/validation/test split of the datasets as Dwivedi et al \yrcite{dwivedi2020benchmarking} and Rampasek et al \yrcite{rampasek2022recipe}.
For each setting, we run the experiment for the same number of times as their work using different random seeds and report the mean and standard deviation. % of the 10 results.
\fi

\iffalse
\begin{itemize}
  \item \textbf{Accuracy}. Weighted average node classification accuracy is used for the node classification task (PATTERN and CLUSTER), and classification accuracy is used  for the graph classification task (MNIST and CIFAR10).
  \item  \textbf{F1 score} for the positive class is used for performance  evaluation on the TSP dataset, due to high class imbalance, \emph{i.e.,} only the edges in the TSP tour are labeled as positive.
  %\item  \textbf{F1 score}.  Due to high class imbalance, \emph{i.e.,} only the edges in the TSP tour are labeled as positive, F1 score .
  \item \textbf{Hits@K} \cite{hu2020open} is used for the  COLLAB dataset, aiming to measure the model's ability to predict future collaboration relationships.
      This method ranks each true collaboration against 100,000 randomly sampled negative collaborations and counts the ratio of positive edges that are ranked at $K$-th place or above.
  \item \textbf{MAE} (mean absolute error) is used to evaluate graph regression performance on ZINC.
\end{itemize}
\fi

\begin{table*}[t]
\caption{Results for superpixel graph classification on MNIST and CIFAR10. We show that our RandAlign consistently improves the performance of the base graph convolutional network models. Residual connection and batch normalization,  which are simple strategies that
can help to alleviate  over-smoothing, are applied to the GCN and GAT base models.}
\label{tab:mnist_cifar10}
\vskip 0.15in
\begin{center}
%\begin{normalsize}
%\begin{sc}
\small
\begin{tabular}{l|c|cccc}
\toprule

%   \multicolumn{6}{c}{MNIST}  \\ % &\multirow{2}{*}{\#GCN layers}
%\cmidrule(lr){1-6}

%Model &Mode &4 layers & 8 layers & 12 layers & 16 layers\\
%\hline
\multirow{2}{*}{Model}   & \multicolumn{5}{c}{MNIST}  \\ % &\multirow{2}{*}{\#GCN layers}
%\cmidrule(lr){2-6}
\cline{2-6}
    &Mode &4 layers & 8 layers & 12 layers & 16 layers\\
\hline
\hline

GCN                &\multirow{2}{*}{ Training} &97.196$\pm$0.223 &99.211$\pm$0.421 &99.862$\pm$0.043 &99.697$\pm$0.029 \\

GCN   + RandAlign  &  &88.311$\pm$0.262 &92.450$\pm$0.170 &94.283$\pm$0.192 &95.505$\pm$0.154  \\
\hline
\rowcolor{LightCyan}
GCN         & &90.705$\pm$0.218 &90.847$\pm$0.078 &91.263$\pm$0.216 &91.147$\pm$0.185 \\
\rowcolor{LightCyan}
\textbf{GCN   + RandAlign}  &\multirow{-2}{*}{ Test}  &\textbf{90.305$\pm$0.140} &\textbf{92.688$\pm$0.046} & \textbf{93.470$\pm$0.035} &\textbf{94.051$\pm$0.052}  \\
\hline
\hline

GAT               &\multirow{2}{*}{Training} &99.994$\pm$0.008 &100.00$\pm$0.000 &100.00$\pm$0.000 &100.00$\pm$0.000 \\
GAT + RandAlign  &  &96.853$\pm$0.236 &98.492$\pm$0.294 &99.146$\pm$0.104 &99.189$\pm$0.158 \\
\hline
\rowcolor{LightCyan}
GAT               & &95.535$\pm$0.205 &96.065$\pm$0.093 &96.288$\pm$0.049 &96.526$\pm$0.041 \\
\rowcolor{LightCyan}
\textbf{GAT + RandAlign}  &\multirow{-2}{*}{Test}  &\textbf{96.513$\pm$0.075} &\textbf{97.250$\pm$0.049} &\textbf{97.505$\pm$0.029} &\textbf{97.553$\pm$0.034} \\

\hline
\hline
%\midrule
GatedGCN &\multirow{2}{*}{ Training} &100.00$\pm$0.000 &100.00$\pm$0.000 &100.00$\pm$0.000 &100.00$\pm$0.000  \\
GatedGCN + RandAlign              &         &99.713$\pm$0.094       &99.933$\pm$0.048   &99.849$\pm$0.020       &99.813$\pm$0.023 \\
\hline
\rowcolor{LightCyan}
GatedGCN & &97.340$\pm$0.143 &97.950$\pm$0.023 &98.108$\pm$0.021 &98.132$\pm$0.022  \\
\rowcolor{LightCyan}
\textbf{GatedGCN + RandAlign} &\multirow{-2}{*}{Test}        &\textbf{98.120$\pm$0.076} &\textbf{98.463$\pm$0.079} &\textbf{98.494$\pm$0.054} &\textbf{98.552$\pm$0.023} \\
\bottomrule
\end{tabular}
%\end{sc}
%\end{normalsize}
\end{center}

\begin{center}
%\begin{normalsize}
%\begin{sc}
\small
\begin{tabular}{l|c|cccc}
\toprule

%   \multicolumn{6}{c}{CIFAR10}  \\ % &\multirow{2}{*}{\#GCN layers}
%\cmidrule(lr){1-6}
%Model &Mode &4 layers & 8 layers & 12 layers & 16 layers\\

\multirow{2}{*}{Model}   & \multicolumn{5}{c}{CIFAR10}  \\ % &\multirow{2}{*}{\#GCN layers}
%\cmidrule(lr){2-6}
\cline{2-6}
    &Mode &4 layers & 8 layers & 12 layers & 16 layers\\
%\hline
%\midrule
\hline
\hline

GCN         &\multirow{2}{*}{ Training} &69.523$\pm$1.948  &77.546$\pm$0.813 &81.073$\pm$1.224 &84.279$\pm$0.656 \\
GCN   + RandAlign  &  &59.798$\pm$0.324 &65.405$\pm$0.603 &66.711$\pm$0.338 &70.919$\pm$0.522   \\
\hline
\rowcolor{LightCyan}
GCN       & &55.710$\pm$0.381  &54.242$\pm$0.454 &53.867$\pm$0.090 &53.353$\pm$0.184 \\
\rowcolor{LightCyan}
\textbf{GCN   + RandAlign}  &\multirow{-2}{*}{Test} &\textbf{55.275$\pm$0.165} &\textbf{57.145$\pm$0.202} &\textbf{57.603$\pm$0.157} &\textbf{57.736$\pm$0.162}   \\
\hline
\hline
GAT               &\multirow{2}{*}{ Training} &89.114$\pm$0.499  &99.561$\pm$0.064 &99.972$\pm$0.005 &99.980$\pm$0.003 \\
GAT + RandAlign  &  &74.522$\pm$1.179 &81.071$\pm$0.596 & 81.511$\pm$0.464 & 79.962$\pm$0.142l  \\
\hline
\rowcolor{LightCyan}
GAT               & &64.223$\pm$0.455  &64.452$\pm$0.303 &64.423$\pm$0.121 &64.340$\pm$0.146 \\
\rowcolor{LightCyan}
\textbf{GAT + RandAlign}  &\multirow{-2}{*}{Test}  &\textbf{65.385$\pm$0.074} &\textbf{69.158$\pm$0.438} &\textbf{69.707$\pm$0.350}   &\textbf{69.920$\pm$0.082} \\

\hline
\hline
%\midrule
GatedGCN &\multirow{2}{*}{ Training} &94.553$\pm$1.018 &99.983$\pm$0.006 &99.995$\pm$0.003 &99.995$\pm$0.004  \\
GatedGCN + RandAlign          &      &77.784$\pm$0.799  &83.552$\pm$0.570 &86.779$\pm$0.520 &90.903$\pm$0.785\\
\hline
\rowcolor{LightCyan}
GatedGCN & &67.312$\pm$0.311 &69.808$\pm$0.421 &68.417$\pm$0.262 &70.007$\pm$0.165  \\
\rowcolor{LightCyan}
\textbf{GatedGCN + RandAlign} &\multirow{-2}{*}{Test}  &\textbf{72.075$\pm$0.154} &\textbf{75.015$\pm$0.177} &\textbf{76.135$\pm$0.248} &\textbf{76.395$\pm$0.186} \\
\bottomrule
\end{tabular}
%\end{sc}
%\end{normalsize}
\end{center}

\end{table*}

\begin{figure*}[ht]
%\vskip 0.2in
\begin{center}
\centerline{\includegraphics[width=0.86\textwidth]{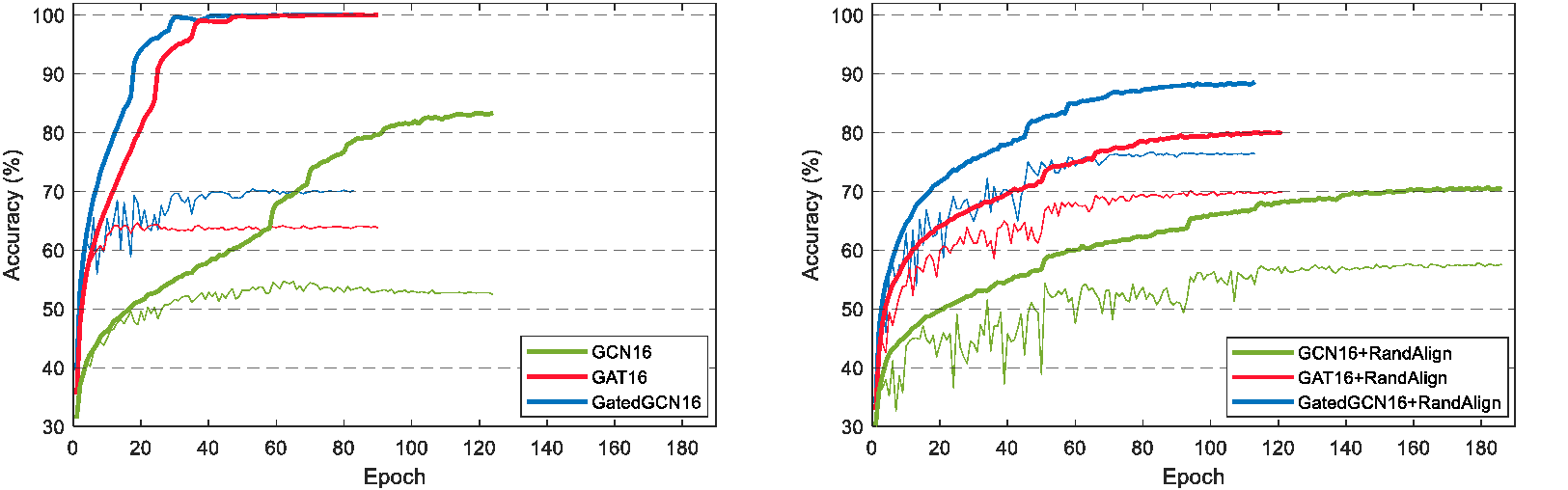}}
\caption{Learning curves on  CIFAR10. Bold lines are training curves and thin lines are test curves. We show that our RandAlign method improves the generalization performance by reducing the issue of over-smoothing.}%We show that the overfitting issue is effectively reduced through tackling the issue of over-smoothing with our Lrn\&Align method.}
\label{fig:learning_curve}
\end{center}
\vskip -0.2in
\end{figure*}

\subsection{Experimental Results}
\textbf{Superpixel Graph Classification on MNIST and CIFAR10}.
The quantitative results  on MNIST and CIFAR10 for superpixel graph classification are reported in Table \ref{tab:mnist_cifar10}.
We experiment with three different base models: GCN, GAT and GatedGCN.
We also applied residual connections \cite{he2016deep} and batch normalizations \cite{ioffe2015batch} to the base models of GCN and GAT.
%Residual connection and batch normalization can reduce the over-smoothing issue and  empirically
Residual connection and batch normalization are simple strategies which are empirically helpful to  reduce the over-smoothing issue and improve the numerical stability in optimization.
GatedGCN employs the gated update approach in aggregating information from neighbours and also integrates residual connections and batch normalizations.
%GatedGCN  is a message-passing-based GNN that employs  edge gates, as well as residual connections, batch normalization, in the model.
We see that the base models  only slightly improve the performance or see a reduced performance as the number of layers increases from 4 to 16.
Without residual connection and batch normalization, the performance would drop considerably with increased layers due to over-smoothing.
By integrating the RandAlign regularization method into the models, the performance of the base models consistently improves as the number of layers  increases.
RandAlign on GatedGCN with 16 layers yields a 6.388\% performance improvement on CIFAR10, which is a 9.13\% relative improvement.
%Our Lrn\&Align on the three models consistently improves the model performance as the layer number increases.
We also see that for the 4 layer GCN model, applying RandAlign could not improve the performance on the two datasets.
This is because the model does not suffer the over-smoothing issue at this layer.

\iffalse
Applying our Lrn\&Align on GatedGCN yields a 6.0\% performance improvement, which is a 28\% relative improvement, on CIFAR10.
We also see that for the 4 layer GCN model, applying Lrn\&Align could not improves the performance.
This is because the model dose not suffer the over-smoothing issue.
\fi

We see from Table \ref{tab:mnist_cifar10} that the base models suffer serious over-fitting problem on the two datasets.
For example, the GAT and GatedGCN with 8 or more graph convolutional layers obtain  nearly 100\% training accuracy on CIFAR10, but their  test accuracy is below  70.007\%.
By using our RandAlign regularization method, we see that all their training accuracy reduces while the task performance improves.
This shows that through tackling the over-smoothing issue with our RandAlign, the over-fitting problem is significantly reduced, and therefore the model generalization performance is improved.
Figure \ref{fig:learning_curve} demonstrates the learning curves of the three base models with 16 layers on the CIFAR10.

\iffalse
We also see that Lrn\&Align reduces  the training accuracy while improving the task performance.
Our results show that through tackling the over-smoothing issue, the model generalization performance  is significant improved.

We note that the performance of GCN improves as layer number increases, this is because normalization and residual connection are applied to the model.

For results on superpixel classification, we first show that our Lrn\&Align method improves the model performance with increase graph convolutional layers.
We apply our Lrn\&Align method on three models: GCN, GAT and GatedGCN.
The GatedGCN architecture is an anisotropic message-passing-based GNN that employs residual connections, batch normalization, and edge gates.
\fi

Table \ref{comp-minst-cifar10} compares the performance of our results with the recent methods on MNIST and CIFAR10.
EGT \cite{hussain2022global}, which integrates an additional edge channels into the Transformer model and also uses  global self-attention to generate embeddings, achieves 98.173\% accuracy on MNIST, which is the best among the previous models.
Our model achieves 0.337\% improved performance compared with EGT \cite{hussain2022global}.
On CIFAR10, our model outperforms the previous best model DGN \cite{beaini2021directional} by 3.557\%, which is a significant improvement.
Our RandAlign also outperforms the SSFG regularization method \cite{zhang2022ssfg}, which essentially stochastically scales features and gradients during training for regularization graph neural network models, but the SSFG method involves a time-consuming parameter tuning process.
To the best of our knowledge, our method achieves the state of the art results on the two datasets.

\begin{table}[tbp]

\caption{Comparison with previous methods  on MNIST and CIFAR10 on superpixel graph classification.}
\label{comp-minst-cifar10}
\vskip 0.15in

\centering  % 表居中
\begin{tabular}{l cccc }
\toprule
%\multirow{2}{*}{Model}   & \multicolumn{2}{c}{Dataset}  \\ % &\multirow{2}{*}{\#GCN layers}
%\cmidrule(lr){2-3}
Model &MNIST &CIFAR10 \\
\midrule

GCN \cite{kipf2016semi}  &90.705$\pm$0.218 &55.710$\pm$0.381 \\

MoNet \cite{monti2017geometric} &90.805$\pm$0.032 &54.655$\pm$0.518 \\
GraphSAGE \cite{hamilton2017inductive} &97.312$\pm$0.097 &65.767$\pm$0.308\\

GIN   \cite{xu2019powerful}     &96.485$\pm$0.252 &55.255$\pm$1.527\\
GCNII \cite{chen2020simple}   &90.667$\pm$0.143   &56.081$\pm$0.198  \\
%GAT       &95.535$\pm$0.205 &64.223$\pm$0.455 \\

%GatedGCN &97.340$\pm$0.143 &67.312$\pm$0.311\\

PNA \cite{corso2020principal}  &97.94$\pm$0.12 &70.35$\pm$0.63\\

DGN   \cite{beaini2021directional}    &-- &72.838$\pm$0.417\\

CRaWl  \cite{toenshoff2021graph}  &97.944$\pm$0.050 &69.013$\pm$0.259 \\
GIN-AK+ \cite{zhao2021stars}  &-- &72.19$\pm$0.13\\
3WLGNN \cite{maron2019provably}  &95.075$\pm$0.961 &59.175$\pm$1.593 \\

EGT  \cite{hussain2022global}     &98.173$\pm$0.087 &68.702$\pm$0.409 \\

GPS \cite{bresson2017residual}      &98.051$\pm$0.126 &72.298$\pm$0.356 \\

GatedGCN + SSFG \cite{zhang2022ssfg} &97.985$\pm$0.032 &71.938$\pm$0.190 \\

\midrule
GAT-16  \cite{velickovic2018graph}     &95.535$\pm$0.205 &64.223$\pm$0.455 \\
\textbf{GAT-16 + RandAlign}   &\textbf{97.553$\pm$0.034} &\textbf{69.920$\pm$0.082}   \\

\hline

% GatedGCN-16 + Lrn\&Align   &98.512$\pm$0.033 &76.395$\pm$0.186   \\
GatedGCN-16 \cite{bresson2017residual} &97.340$\pm$0.143 &67.312$\pm$0.311\\

%\textbf{GatedGCN-16 +}   &\multirow{2}{*}{\textbf{98.512$\pm$0.033}} %&\multirow{2}{*}{\textbf{76.395$\pm$0.186}}   \\
\textbf{GatedGCN-16 + RandAlign}  &\textbf{98.512$\pm$0.033} &\textbf{76.395$\pm$0.186}\\  %\hspace{15pt}  \hskip 0.25in
\bottomrule

\end{tabular}

\label{table:results_graphclassification}
\end{table}

\begin{table}[tbp]

\caption{Comparison with previous work on PascalVOC-SP on the superpixel graph classification task.}
\label{comp-voc}
\vskip 0.15in

\small
\centering  % 表居中
\begin{tabular}{l cc }
\toprule
%\multirow{2}{*}{Model}   & \multicolumn{2}{c}{Dataset}  \\ % &\multirow{2}{*}{\#GCN layers}
%\cmidrule(lr){2-3}
Model &PascalVOC-SP \\
\midrule

GCN \cite{kipf2016semi} & 0.1268 $\pm$ 0.0060  \\

GINE \cite{hu2019strategies}  & 0.1265 $\pm$ 0.0076  \\

GatedGCN \cite{bresson2017residual} &0.2873 $\pm$ 0.0219  \\

GatedGCN + RWSE \cite{dwivedi2021graph}  &0.2860 $\pm$ 0.0085  \\

Transformer + LapPE \cite{dwivedi2022long}  &0.2694 $\pm$ 0.0098  \\

SAN + LapPE \cite{dwivedi2022long}  &0.3230 $\pm$ 0.0039  \\

SAN + RWSE \cite{dwivedi2022long}  &0.3216 $\pm$ 0.0027  \\

\hline

GPS \cite{rampasek2022recipe} &0.3748 $\pm$ 0.0109  \\
\textbf{GPS + RandAlign}   &0.4288 $\pm$ 0.0062 \\  %($K$=8)

\bottomrule

\end{tabular}

\label{table:results_cocovoc}
\end{table}

\begin{table}[tbp]
\caption{Experimental results  PATTERN and CLUSTER on the node classification task.}
\label{table:cluster}
\vskip 0.15in
\centering  % 表居中
\small

\centering  % 表居中
\begin{tabular}{lccccc }
\toprule[1pt]

%\multirow{2}{*}{Model}   & \multicolumn{2}{c}{CLUSTER}  \\ % &\multirow{2}{*}{\#GCN layers}
%\cmidrule(lr){2-3}
Model &PATTERN  &CLUSTER \\
\midrule

GCN \cite{kipf2016semi}
       &71.892$\pm$0.334 &68.498$\pm$0.976\\
GraphSAGE \cite{hamilton2017inductive}
                                             &50.492$\pm$0.001 &63.844$\pm$0.110 \\
GIN \cite{xu2019powerful}
                              &85.387$\pm$0.136 &64.716$\pm$1.553\\

GAT \cite{velickovic2018graph}
                                        &78.271$\pm$0.186 &70.587$\pm$0.447\\
RingGNN \cite{chen2019equivalence}   & 86.245$\pm$0.013   &42.418$\pm$20.063\\

MoNet \cite{monti2017geometric}
                                 &85.582$\pm$0.038 &66.407$\pm$0.540\\

GatedGCN \cite{bresson2017residual}
                                         &85.568$\pm$0.088 &73.840$\pm$0.326\\

DGN \cite{beaini2021directional}  &86.680 $\pm$ 0.034 & --\\
%SAN \cite{kreuzer2021rethinking} & &76.691$\pm$0.65 \\
K-Subgraph SAT \cite{chen2022structure}  &86.848$\pm$0.037 &77.856$\pm$0.104 \\
GatedGCN + SSFG \cite{zhang2022ssfg} &85.723$\pm$0.069 &75.960$\pm$0.020 \\
\midrule
SAN \cite{kreuzer2021rethinking}  &86.581$\pm$0.037 &76.691$\pm$0.650 \\
\textbf{SAN + RandAlign} &\textbf{86.770$\pm$0.067}  &\textbf{77.847$\pm$0.073} \\
\hline
GPS \cite{rampasek2022recipe} &86.685$\pm$0.059 & 78.016$\pm$0.180\\
\textbf{GPS + RandAlign} &\textbf{86.858$\pm$0.010} &\textbf{78.592$\pm$0.052} \\
\bottomrule
\end{tabular}

\end{table}

\iffalse
EGT incorporates an edge channels into the Transformer and uses the global self-attention to learn embeddings.
On CIFAR10, our model outperforms the previous best DGN \cite{beaini2021directional} by 3.557\%, which is a large improvement.
Our method also outperforms SSFT, but SSFG does not constrain the smoothness among node embeddings.
Our results also outperform GPS, which uses a message passing and transformer to generate embeddings.
To the best of our knowledge, our graph convolutional network achieves the state of the art performance on the two datasets.
\fi

\textbf{Results on PascalVOC-SP}.
PascalVOC-SP is a long range superpixel classification dataset as compared to MNIST and CIFAR10.
Table \ref{comp-voc} reports the results  on  this dataset. %PascalVOC-SP on superpixel classification.
We experiment with GPS \cite{rampasek2022recipe} as the base model.
The GPS  model  uses a graph convolutional network and a Transformer to model local and global dependencies in the graph.
This model archives the best performance among the baseline models.
We see from Table \ref{comp-voc} that our RandAlign improves the performance of GPS from 37.48\% to 42.88\%, which is a 14.41\% relative improvement.
Once again, our RandAlign method improves the performance of the base model, advancing the state of the art result for long range graph representation learning on this dataset.

\textbf{Node Classification on PATTERN and CLUSTER}.
%Table \ref{table:pattern} and Table \ref{table:cluster} report our results for node classification on PATTERN and CLUSTER.
Table \ref{table:cluster} reports the experimental results on PATTERN and CLUSTER on node classification.
We experiment with two state of the art base
architectures: the spectral attention network (SAN) \cite{kreuzer2021rethinking}  and GPS \cite{rampasek2022recipe}.
%We use SAN \cite{kreuzer2021rethinking} and GPS \cite{rampasek2022GPS} as the baseline models.
SAN  utilizes an invariant aggregation of Laplacian's eigenvectors for position encoding and also utilizes conditional attention for the real and virtual edges to improve the performance. %Recipe for a General, Powerful, Scalable Graph Transformer
As introduced above, a GPS layer integrates a message passing graph convolutional layer  and a Transformer layer to learn local and global dependencies.
%A GPS layer is a hybrid message passing network and Transformer layer.
We see from Table \ref{table:cluster} our RandAlign regularization method improves the performance of the two base models and advances the state of the results on the two datasets.
It improves the performance by 1.156\% on SAN and 0.576\% on GPS on the CLUSTER dataset.
Our model achieves considerably improved performance when compared with GCN, GAT and GraphSAGE.
Notably, the GPS model with  RandAlign regularization outperforms all the baseline models on the two datasets.

\textbf{Multi-label Graph Classification on Peptides-func}.
Table \ref{table:Peptides-struct} reports the results on Peptides-func.
This dataset was introduced to evaluate a model's ability to capture long-range dependencies in the  graph.
We also experiment with  GPS \cite{rampasek2022recipe} as the base model.
%We followed the same experimental setup and implementation details as \cite{rampasek2022GPS} and  applied our Lrn\&Align on the GPS model.
As aforementioned,  the GPS model combines a Transformer layer with the message passing graph convolutional network framework to capture the global dependencies.
%The GPS model achieves the best result among the baseline models on Peptides-func.
We see from Table \ref{table:Peptides-struct} that our RandAlign improves the average precision of GPS from 0.6535 to 0.6630,  outperforming all the baseline models including GatedGCN, Transformer \cite{vaswani2017attention} and SAN.
Peptides-struct is also a long range graph dataset, as with the PascalVOC-SP dataset.
The results on the two datasets also  show that RandAlign helps to improve the  performance for capturing long-range dependencies in the graph in graph representation learning.

%GPS fully connected Graph Transformers [60, 32, 41] are able to model long-range dependencies in the graphs.

\begin{table}[tbp]

\caption{Experimental results on Peptides-func on the multi-label graph
classification task.}
\label{table:Peptides-struct}
\vskip 0.15in
\centering  % 表居中

\small
\begin{tabular}{lccccc }
\toprule[1pt]

%Dataset  &$\#$Graphs & $\#$Nodes &$\#$training &$\#$Val. &$\#$Test \\
%\multirow{2}{*}{ Model}    & \multicolumn{1}{c}{Peptides-func}  \\
%\cmidrule(lr){2-2}
Model &  AP ($\uparrow$)  \\
\midrule[.6pt]

GCN \cite{kipf2016semi}  &0.5930$\pm$0.0023\\

GINE \cite{hu2019strategies}         &0.5498$\pm$0.0079  \\

GatedGCN  \cite{bresson2017residual}       &0.5864$\pm$0.0077 \\

GatedGCN + RWSE \cite{dwivedi2021graph}   &0.6069$\pm$0.0035\\

Transformer + LapPE \cite{dwivedi2022long}  &0.6326$\pm$0.0126 \\
%\midrule[.5pt]
%Transformer  \cite{vaswani2017attention} &\multirow{2}{*}{0.6326$\pm$0.0126} \\
%\hspace{6pt} + LapPE & &\\
SAN  + LapPE \cite{dwivedi2022long}   &0.6384$\pm$0.0121\\
SAN   + RWSE \cite{dwivedi2022long} &0.6439$\pm$0.0075 \\
\midrule[.5pt]

GPS \cite{rampasek2022recipe} &0.6535$\pm$0.0041\\
\textbf{GPS   + RandAlign}  &\textbf{0.6630$\pm$0.0005} \\

\bottomrule[1pt]

\end{tabular}

\label{table:peptides-struct}
\end{table}

\begin{table}[tbp]

\caption{Experimental results on OGBG-molhiv on binary graph classification. The models are all trained from scratch.}
\vskip 0.15in
\centering  % 表居中
%\scriptsize
\small
\begin{tabular}{lccccc }
\toprule[1pt]

%Dataset  &$\#$Graphs & $\#$Nodes &$\#$training &$\#$Val. &$\#$Test \\
%\multirow{2}{*}{Model}    & \multicolumn{1}{c}{ogbg-molhiv}  \\
%\cmidrule(lr){2-2}
Model &  ROC-AUC ($\uparrow$)   \\
\midrule[.6pt]

GCN  \cite{kipf2016semi}   &0.7599$\pm$0.0119 \\

GIN \cite{xu2019powerful}         &0.7707$\pm$0.0149  \\

PNA \cite{corso2020principal}         &0.7905$\pm$0.0132 \\
DeeperGCN  \cite{li2020deepergcn}  &0.7858$\pm$0.0117 \\
DGN  \cite{beaini2021directional} & 0.7970$\pm$0.0097 \\

ExpC \cite{yang2022breaking} &0.7799$\pm$0.0082 \\
GIN-AK+ \cite{zhao2022from} &0.7961$\pm$0.0119 \\

SAN  \cite{kreuzer2021rethinking} &0.7785$\pm$0.2470 \\
\midrule[.5pt]

GPS \cite{rampasek2022recipe}  &0.7880$\pm$0.0101\\
\textbf{GPS + RandAlign}  &\textbf{0.8021$\pm$0.0305} \\

\bottomrule[1pt]

\end{tabular}

\label{table:ogbg-molhiv}
\end{table}

\begin{table}[tbp]

\caption{Importance of scaling embeddings of the previous layer  in  alignment.}
\label{comp:lambda_uniform}
\vskip 0.15in

\small
\centering  % 表居中
\begin{tabular}{l cccc }
\toprule
%\multirow{2}{*}{Method}   & \multicolumn{2}{c}{Dataset}  \\ % &\multirow{2}{*}{\#GCN layers}
%\cmidrule(lr){2-3}
Model &MNIST &CIFAR10 \\
\midrule[.75pt]

GAT-8\\
\hskip 0.1in w/o Lrn\&Align   &96.065$\pm$0.093 &64.452$\pm$0.303 \\

\hline
\hskip 0.1in \textbf{RandAlign w/o scaling}  &\textbf{96.977$\pm$0.021}  & \textbf{66.212$\pm$0.182} \\

\hskip 0.1in RandAlign + scaling  &97.250$\pm$0.049 &69.158$\pm$0.438   \\

\midrule[.75pt]
GatedGCN-8 \\
\hskip 0.1in w/o RandAlign  &97.950$\pm$0.023 &69.808$\pm$0.421 \\
\hline

\hskip 0.1in \textbf{RandAlign w/o scaling}   &\textbf{98.247$\pm$0.018}  &\textbf{74.437$\pm$0.150}  \\

\hskip 0.1in RandAlign + scaling  &98.463$\pm$0.079 &75.015$\pm$0.177  \\

\bottomrule

\end{tabular}

\label{table:results_graphclassification}
\end{table}

\textbf{Binary Graph Classification on OGBG-molhiv}.
The results on OGBG-molhiv are reported in Table \ref{table:ogbg-molhiv}.
As with Rampasek et al.  \cite{rampasek2022recipe}, we only compare with the baseline models that are trained from scratch.
We experiment using  GPS as the base model.
It can be seen that RandAlign improves the ROC-AUC of GPS from 0.6535 to 0.6630, which is a relative 1.45\% improvement, outperforming all the baseline models, including PNA \cite{corso2020principal}, DGN  \cite{beaini2021directional} and GIN-AK+ \cite{zhao2022from}.

We have shown that RandAlign is a general method for preventing the over-smoothing issue.
It improves the generalization performance of different graph convolutional network models and on different domain tasks.
We also see from the experimental results that applying RandAlign results in a small standard deviation for most experiments compared with the base models.
This suggests that RandAlign is also effective for improving  numerical stability when optimizing the graph convolutional network models.

\iffalse
\textbf{Tackling the over-smoothing issue}.
Over-smoothing is a common issue with graph convolutional networks.
%The over-smoothing issue comes after repeatedly applying  graph convolutional layers, the representations of the nodes converge to similar values.
The over-smoothing issue comes with repeatedly applying  graph convolutions, resulting in node features converging to similar values.
Because of the over-smoothing issue, most graph convolutional networks are restricted to shallow layers, \emph{e.g.,} 2 to 4.
Further increasing the number of graph convolutional layers will lead to reduced performance.

Figure \ref{fig:accuracy} shows the training and test accuracies/F1 scores/MAEs of our model with different graph convolutional layers on the benchmark datasets.
We see that our training and test performances improve consistently as the number of graph convolutional layers increases.
This suggests that our two-stage local feature aggregation method helps to address the over-smoothing issue.
\fi

\textbf{Importance of Scaling $\mathbf{h}_u^{(k-1)}$ in Alignment}.
In our RandAlign method, we first scale $\mathbf{h}_u^{(k-1)}$ to have the norm of  $\frac{\mathbf{h}_u^{(k-1)}}{\lVert \mathbf{h}_u^{(k-1)}\rVert}  {\lVert \overline{\mathbf{h}}}_v^{(k)}\rVert$ and then apply a random interpolation between the scaled feature and $\overline{\mathbf{h}}_{v}^{(k)}$  (see Equation \ref{eq:rand_alignment}) for aligning $\overline{\mathbf{h}}_{v}^{(k)}$.
To show the importance of the scaling step, we further validate our method without the scaling step.
The experiments are carried out on MNIST and CIFAR10 using  GAT-8 and  GatedGCN-8 as the base models, and Table \ref{comp:lambda_uniform} reports the comparison results.
We see that applying scaling improves the performance of the two base models on the two datasets.
By scaling $\mathbf{h}_u^{(k-1)}$ to have the  norm of $\overline{\mathbf{h}}_v^{(k)}$, more information about $\overline{\mathbf{h}}_v^{(k)}$ is contained in the aligned representation, and therefore  the task performance is improved.

\section{Conclusions}
Over-smoothing is a common issue in message-passing  graph convolutional networks.
In this paper, we proposed RandAlign for regularizing graph convolutional networks through reducing the over-smoothing problem.
The basic idea of RandAlign is to randomly align the generate embedding for each node and with that generated by the previous layer in each message passing iteration. % by reducing the angle between these embeddings.
Our method is motivated by the intuition that  learned  embeddings for the nodes  become smoothed layerwisely or asymptotically layerwisely.
%We provided an illustration example to show this intuition.
In our RandAlign, a random interpolation method is utilized for feature alignment.
By aligning the generated embedding for each node with that generated by the previous layer, the smoothness of these embeddings is reduced.
Moreover,  we introduced a scaling step to scale the embedding of the previous layer to the same
norm as the generated embedding before performing random interpolation.
This scaling step can better maintain the benefit yielded by graph convolution in the aligned embeddings.
The proposed RandAlign is a parameter-free method, and it can be directly applied current graph convolutional networks without introducing additional trainable weights and the hyper-parameter tuning procedure.
We experimentally evaluated RandAlign on seven popular benchmark datasets on four graph domain tasks including graph classification, node classification, multi-label graph classification and binary graph classification.
We presented extensive results to demonstrate RandAlign is a generic method that improves the performance of a variety of graph convolutional network models and advances the state of the art results for graph representation learning.

\section{Acknowledgement}
We would like to  thank the editor and anonymous reviewers for reviewing our manuscript.

%\clearpage
\bibliographystyle{IEEEtran}
\bibliography{myref}

\end{document}